\documentclass[conference]{IEEEtran}
\usepackage{multicol}
\usepackage{multirow}
\usepackage{lipsum}
\usepackage{booktabs}
\usepackage{amsfonts}
\usepackage{amsmath,cases}
\usepackage{amssymb}
\usepackage{graphicx}
\usepackage{cite}
\usepackage{epsfig}
\usepackage[T1]{fontenc}
\usepackage[lowtilde]{url}
\usepackage{color}
\usepackage{algorithm}
\usepackage{algorithmic}
\usepackage{epstopdf}
\usepackage{color}
\usepackage [english]{babel}
\usepackage [autostyle, english = american]{csquotes}
\MakeOuterQuote{"}
\usepackage{color}
\usepackage{amssymb}
\usepackage{bm}
\usepackage{caption}
\usepackage{subcaption}

\usepackage{float}
\usepackage{multirow}
\usepackage{hhline}

\usepackage{algorithm,algorithmic}
\usepackage{array}
\ifCLASSOPTIONcompsoc
\usepackage[caption=false,font=footnotesize,labelfont=sf,textfont=sf]{subfig}
\else
\usepackage[caption=false,font=footnotesize]{subfig}
\fi
\usepackage{fixltx2e}
\usepackage{amsthm}
\theoremstyle{plain}

\newcommand{\mc}{\mathcal}
\newcommand{\mb}{\mathbb}

\newcommand{\tb}{\textbf}

\newcommand{\bea}{\begin{eqnarray}}
\newcommand{\eea}{\end{eqnarray}}
\newcommand{\beq}{\begin{equation}}
\newcommand{\eeq}{\end{equation}}

\newtheorem{ex}{Example}\newcommand{\Ex}{\begin{ex}\rm}
\newcommand{\eex}{\end{ex}}

\begin{document}
\title{Concatenated image completion via tensor augmentation and completion}

\author{\IEEEauthorblockN{Johann A. Bengua,  Hoang D. Tuan and Ho N. Phien}
\IEEEauthorblockA{Faculty of Engineering and Information Technology\\
University of Technology Sydney\\
Ultimo, Australia\\
Email: johann.a.bengua@student.uts.edu.au\\
tuan.hoang@uts.edu.au\\
ngocphien.ho@uts.edu.au
}
\and
\IEEEauthorblockN{Minh N. Do}
\IEEEauthorblockA{Department of Electrical and Computer Engineering\\ and the Coordinated Science Laboratory\\
University of Illinois at Urbana-Champaign\\
Illinois, USA\\
Email: minhdo@illinois.edu
}}

\maketitle
\begin{abstract}
This paper proposes a novel framework called concatenated image completion via tensor augmentation and completion (ICTAC), which recovers missing entries of color images with high accuracy. Typical images are second- or third-order tensors (2D/3D) depending if they are grayscale or color, hence tensor completion algorithms are ideal for their recovery. The proposed framework performs image completion by concatenating copies of a single image that has missing entries into a third-order tensor, applying a dimensionality augmentation technique to the tensor, utilizing a tensor completion algorithm for recovering its missing entries, and finally extracting the recovered image from the tensor. The solution relies on two key components that have been recently proposed to take advantage of the tensor train (TT) rank: A tensor augmentation tool called ket augmentation (KA) that represents a low-order tensor by a higher-order tensor, and the algorithm tensor completion by parallel matrix factorization via tensor train (TMac-TT), which has been demonstrated to outperform state-of-the-art tensor completion algorithms. Simulation results for color image recovery show the clear advantage of our framework against current state-of-the-art tensor completion algorithms.
\end{abstract}
\IEEEpeerreviewmaketitle
\section{Introduction}
\begin{figure}[!htbp]
\centering
\includegraphics[width=2.5in]{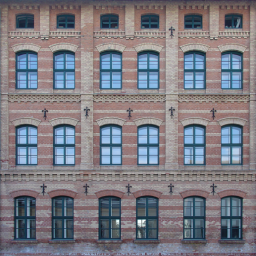}
\caption{Facade color image exhibiting many structural properties such as repetition, symmetry and continuity that can be used for completion.}\label{facexample}
\end{figure}

The problem of recovering missing entries from an array of data is known as completion \cite{Candes_2009}. Often recovery of the missing data will be based on the available elements of the array and its structural properties, but in the event that there is no pattern between elements, then completion would not be possible. Real-world data tends to exhibit correlations, symmetry, repetition and continuity that make completion possible and reliable. The example in Fig. \ref{facexample} demonstrates a multidimensional array, with entries representing red, green and blue (RGB) values to form the color image.
Traditional methods of matrix factorization \cite{Wen2012} and nuclear norm (matrix trace norm) minimization \cite{doi:10.1137/080738970} can be used to solve matrix completion problems. There has been particular success in applying these methods to many fields of research such as compressed sensing \cite{6867345}, magnetic resonance imaging (MRI) \cite{MRM:MRM24997}, gene expression prediction \cite{Kapur2016} and image classification \cite{7083747}.

Algorithms that can complete missing entries of tensors have attracted interest in recent years because of the flexibility and potential applications they offer. Tensors are multidimensional arrays that are higher-order generalizations of matrices and vectors. According to tensor nomenclature \cite{Tamara_2009}, a vector and matrix are considered first-order and second-order tensors, where the order represents the number of dimensions (also known as ways or modes). Some examples of tensors include a color image, which is a third-order tensor defined by two indices for spatial variables and one index for color mode. A video comprised of color images is a fourth-order tensor with an additional index for a temporal variable. The utilization of tensors and their decompositions can be seen in seminal works on computer vision \cite{Vasilescu_2003} and data mining \cite{Sun_2005,Franz_2009}. In \cite{Xu}, the authors demonstrated the superiority of tensor completion algorithms to matrix completion algorithms for completion tasks on hyperspectral images, grayscale/color video and brain MRI images. Their work involves the calculation of multiple low rank matrix factorizations for each mode using an unbalanced matricization scheme (one mode versus the rest), which is based on Tucker rank \cite{Ji2013,Gandy_2011}. Alternatively, \cite{Ji2013} utilized Tucker rank for matrix nuclear norm minimization on each mode to solve relaxed optimization problems for tensor completion. Specifically, they introduced three tensor completion algorithms SiLRTC, FaLRTC and HaLRTC, which are presently used as baselines for benchmarking new tensor completion algorithms.

Our previous work \cite{bengua16} had shown that methods based on Tucker rank are not ideal for completion due to the underlying use of an unbalanced matricization scheme. It was shown using von Neumann entropy that tensor completion based on tensor train (TT) rank \cite{PhysRevLett.91.147902,Oseledets_2011} is more ideal for completion problems due to a balanced matricization scheme, i.e. matricization of a tensor along permutations of modes rather than a single mode, which is the foundation of the TT rank. Historically, TT rank has been predominantly used in physics for simulating quantum many-body systems \cite{Vidal_2004}, with only recent real-world applications of TT rank-based methods in machine learning \cite{7207289,NOVIK2015,ZHAO16} and tensor completion \cite{bengua16}. The proposed TT rank-based algorithm TMac-TT in \cite{bengua16}  outperformed SiLRTC and many state-of-the-art tensor completion algorithms such as FBCP \cite{7010937} and STDC \cite{6587455} in both color image and video recovery problems. The advantage of TMac-TT is attributed to the utilization of a novel preprocessing tensor augmentation scheme known as ket augmentation (KA) that creates a structured block addressing of a tensor, which is advantageous only for TT rank-based methods. However, the disadvantage of using KA directly on an image is that block-artifacts \cite{1212659} are created due to the TT rank optimization from TMac-TT. This effect is minimized for the results on color video completion because the initial fourth-order tensor to complete is reshaped to a third-order tensor by combining the row and temporal indices. This provides different structural properties than completing each frame of the video individually, and gives potential for new patterns to assist in completing the missing entries.

In this paper we address the problem of block-artifacts caused by the tensor augmentation of a color image. A novel framework for image completion is introduced that firstly concatenates copies of an image containing missing elements into a third-order tensor, which is inspired by color video completion in \cite{bengua16}. Then, KA is applied on the tensor, followed by the TMac-TT algorithm for tensor completion, and lastly the recovered image is extracted from the tensor.  For the remainder of the paper we refer to this framework as concatenated Image Completion via Tensor Augmentation and Completion (ICTAC).  Experimental results demonstrate the clear advantage of this framework for image completion problems and we compare the results to KA+TMac-TT (KA combined with TMac-TT) \cite{bengua16} and a recently proposed state-of-the-art algorithm smooth PARAFAC decomposition with quadratic variation (SPC-QV) \cite{YOKO15}, which has been shown to also outperform FBCP and STDC.

The paper is structured as follows. Section \ref{revexst} briefly recalls the existing methods of KA and TMac-TT, which are two components of the proposed ICTAC framework. Section \ref{newfw} outlines in detail the proposed ICTAC framework for image completion. In Section \ref{experres}, color image completion problems are addressed by ICTAC, KA+TMac-TT and SPC-QV. Lastly, Section \ref{conclusend} concludes the paper.

Most of the notations used in the paper are described here and adopted from \cite{Tamara_2009}. A \textit{tensor} is a multidimensional array and its \textit{order}, \textit{way} or \emph{mode} is the number of dimensions it contains. Scalars are zero-order tensors denoted by a lowercase letter, e.g. $x$. Vectors are denoted by a boldface lowercase letter such as \textbf{x}, and matrices are denoted as capital letters, e.g. $X$. A higher-order tensor (third-order or higher) is denoted by calligraphic letters, e.g. $\mc{X}$. We denote $\mc{X}\in\mathbb{R}^{I_1\times I_2\times\cdots\times I_N}$ as a \textit{N}th-order tensor with $I_k$ ($k=1,\ldots, N$) representing the size of the $k$th mode. The elements of $\mc{X}$ are represented by $x_{i_1\cdots i_k\cdots i_N}$, where $1\leq i_k\leq I_k$, $k=1,\ldots, N$. The Frobenius norm of $\mc{X}$ is defined by $||\mc{X}||_F = \sqrt{\sum_{i_1}\sum_{i_2}\cdots\sum_{i_N}x^2_{i_1i_2\cdots i_{N}}}$.

The mode-$n$ fiber of a tensor $\mc{X}\in\mathbb{R}^{I_1\times I_2\times\cdots\times I_N}$ is denoted by \text{\bf x}$_{i_1\ldots i_{n-1}:i_{n+1}\ldots i_{N}}$, which is a vector defined by fixing all indices but $i_n$.

Mode-$n$ matricization (also known as mode-$n$ unfolding or flattening) of a tensor $\mc{X}\in\mathbb{R}^{I_1\times I_2\times\cdots\times I_N}$ is the process of unfolding or reshaping the tensor into a matrix $X_{(n)}\in\mathbb{R}^{I_n\times (I_1\cdots I_{k-1}I_{k+1}\cdots I_N)}$ by rearranging the mode-$n$ fibers to be the columns of the resulting matrix.  Tensor element $(i_1,\ldots, i_{n-1},i_n,i_{n+1},\ldots, i_{N})$ maps to matrix element $(i_n,j)$ such that
\bea
j=1+\sum_{k=1,k\neq n}^{N}(i_k-1)J_k~~\text{with}~~J_k=\prod_{m=1, m\neq n}^{k-1}I_m.
\label{indexj}
\eea
This is an unbalanced matricization scheme because only a single mode represent the matrix row in $X_{(n)}$.

{Mode-$(1,2,\ldots, k)$ matricization} of a tensor $\mc{X}\in\mathbb{R}^{I_1\times I_2\times\cdots\times I_N}$ \cite{Oseledets_2011} is defined as $X_{[k]}\in\mb{R}^{m\times n}$ ($m = \prod_{l=1}^{k}I_l, n=\prod_{l=k+1}^{N}I_l$). The TT rank is defined as the vector $\tb{r} = (r_1, r_2,\ldots, r_{N-1})$, where $r_k$ is the matrix rank of $X_{[k]}$.

\section{Ket augmentation and TMac-TT}\label{revexst}
In this section we review ket augmentation (KA) and TMac-TT, which are two components of the ICTAC framework that will be introduced in the next section.

\begin{figure}[!htbp]
\centering
\includegraphics[width=3.3in]{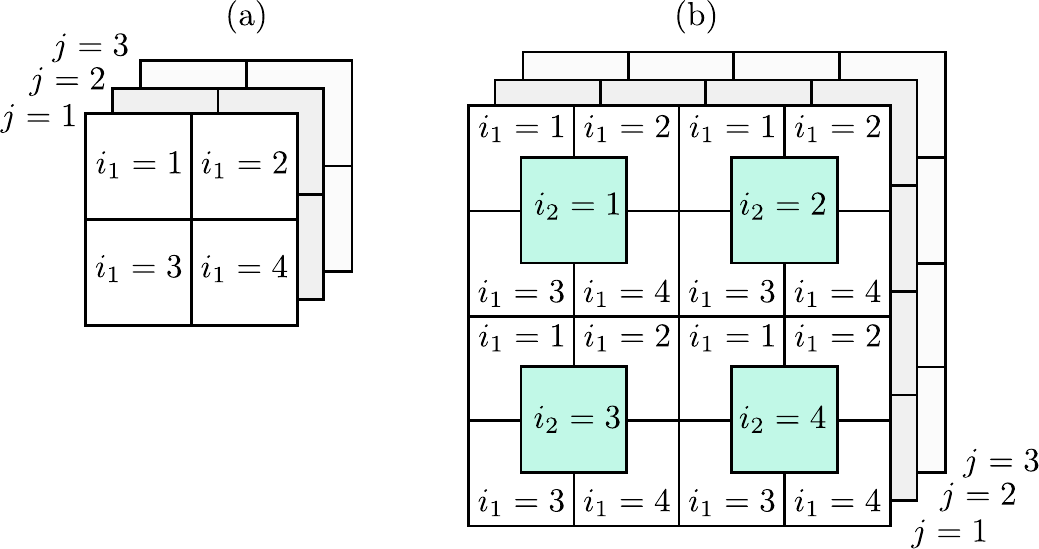}
\caption{A structured block addressing procedure to cast color images into a higher-order tensor. (a) Example addressing for a $2\times2\times3$ color image. (b) Example addressing for a $2^2\times2^2\times3$ color image.}\label{im2ket}
\end{figure}

KA is a tensor augmentation technique that essentially represents a third-order color image tensor
$\mc{T}\in\mathbb{R}^{I_1\times I_2\times I_3}$ ($I_3=3$) into a $K$th-order tensor
$\tilde{\mc{T}}\in\mathbb{R}^{J_1\times J_2\times\cdots\times J_K}$, where $K\geq 3$, $\prod_{l=1}^{3}I_l=\prod_{l=1}^{K}J_l$, and $J_l$ represents a unique block structured addressing of the original image. The general form of $\tilde{\mc{T}}$ is given by
\beq
\tilde{\mc{T}}_{[2^n\times 2^n\times 3]} = \sum_{i_n,\ldots,i_1=1}^{4}\sum_{j=1}^{3}c_{i_n\cdots i_1j}\tb{e}_{i_n}\otimes\cdots\otimes\tb{e}_{i_1}\otimes \tb{u}_{j}.\label{origka2}
\eeq
Fig. \ref{im2ket} demonstrates the result of KA for color two images.
It can be seen that each mode represents different layers of the image, hence by utilizing a TT rank-based algorithm, the correlations of textures can be studied between the modes of the tensor. For this paper we modify the KA algorithm such that
\beq
\tilde{\mc{T}}_{[3^n\times 2^n\times 3]} = \sum_{i_n,\ldots,i_1=1}^{6}\sum_{j=1}^{3}c_{i_n\cdots i_1j}\tb{e}_{i_n}\otimes\cdots\otimes\tb{e}_{i_1}\otimes \tb{u}_{j},\label{moddedKA}
\eeq
where each mode is $i_n=1,\ldots,6$, rather than $i_n=1,\ldots,4$ in (\ref{origka2}). This modification is used for image concatenation, which produces a tensor with one mode significantly larger than the others, e.g. $I_1\gg I_2$ for $\mc{T}$, hence the modified KA in (\ref{moddedKA}) caters for rectangular matrices in the subspace $I_1\times I_2$.

TMac-TT is a tensor completion algorithm that completes a tensor $\mc{X}\in\mathbb{R}^{I_1\times I_2\times\cdots\times I_N}$ containing only a subset of known elements $\Omega$. Specifically, it addresses the following multilinear matrix factorization problem:
\bea
\begin{aligned}
	& \underset{U_k,V_k,X_{[k]}}{\text{min}}&&||U_kV_k-X_{[k]}||^{2}_{F},
\end{aligned}
\label{minrankX7_1}
\eea
for $k=1,2,\ldots,N-1$.  This problem is convex when each variable $U_k,V_k$ and $X_{[k]}$ is optimized while keeping the other two fixed. Specifically, the following steps are performed to update each variable:
\bea
U^{l+1}_{k} &=& X^{l}_{[k]}(V^{l}_{k})^{T},\\
V^{l+1}_{k} &=&( (U^{l+1}_{k})^{T}U^{l+1}_{k})^{\dagger}(U^{l+1}_{k})^{T})X^{l}_{[k]}\\
X^{l+1}_{[k]}&=& U^{l+1}_{k}V^{l+1}_{k},
\label{X1}
\eea
where ``$^\dagger$'' denotes the Moore-Penrose pseudoinverse. 

\begin{figure}[!htbp]
\centering
\includegraphics[width=2.7in]{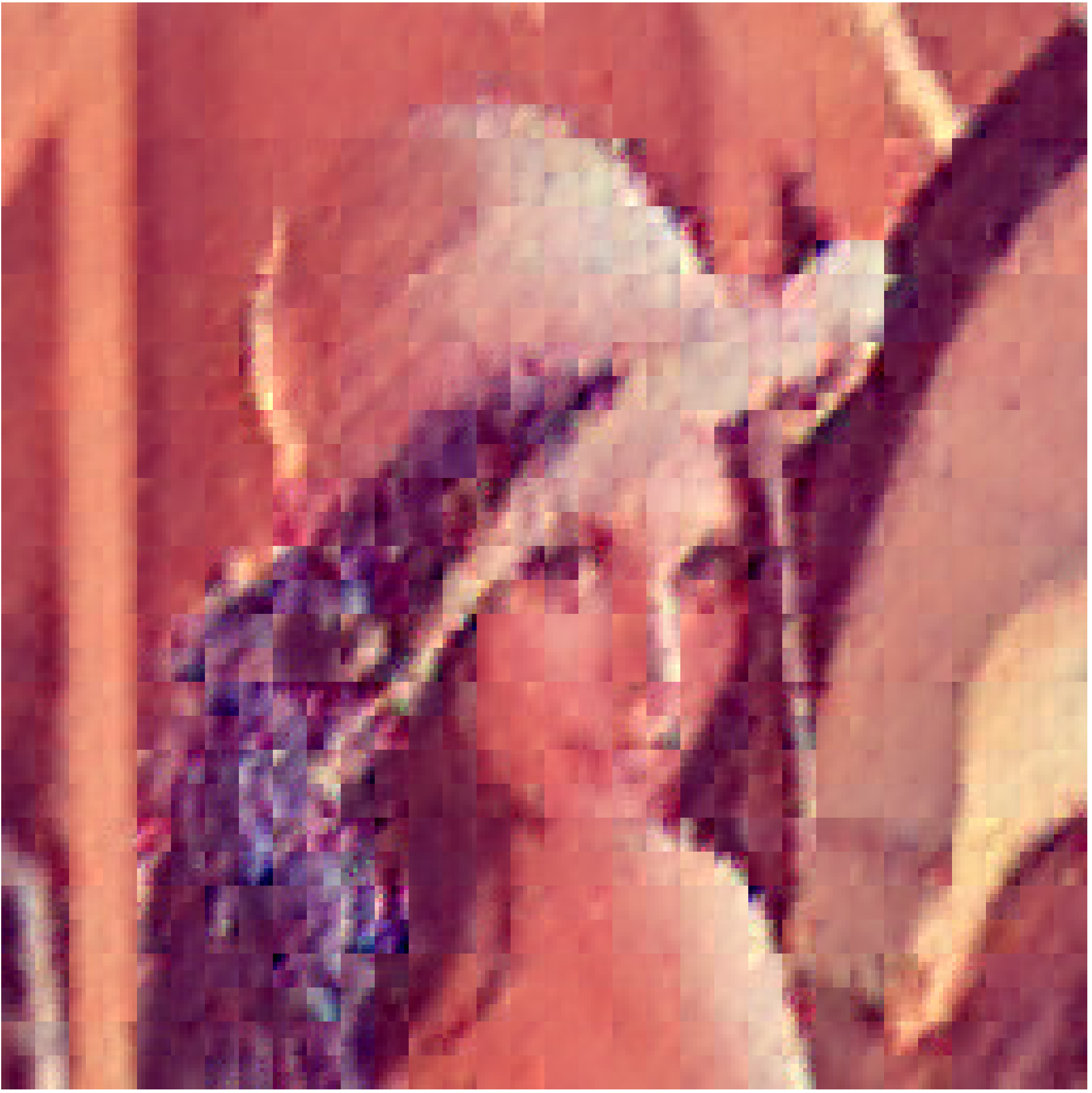}
\caption{Completed Lena image using KA+TMac-TT that previously had 90\% missing entries.}\label{lenablocking}
\end{figure}

It was shown in our previous paper \cite{bengua16} that KA+TMac-TT outperformed baseline algorithms SiLRTC and TMac \cite{Xu}, state-of-the-art algorithms FBCP and STDC, and another TT rank-based algorithm ALS \cite{Holtz_2012,Grasedyck_2015} in color image and video recovery tasks. The next section will introduce a new framework for improving image recovery tasks to minimize block-artifacts caused by the KA on a single image. These artifacts can be seen in the result for completion of the Lena image with 90\% missing entries in Fig. \ref{lenablocking}.
\section{A concatenated image completion framework}\label{newfw}
The concatenated image completion via tensor augmentation and completion (ICTAC) framework is outlined in this section. The framework is divided into three main steps:
\begin{enumerate}
\item\emph{Image concatenation}: The concatenation of a single image with missing entries into a third-order tensor to prepare it for KA, as discussed in Subsection \ref{cifta}.
\item\emph{KA+TMac-TT}: The application of KA then TMac-TT on the concatenated third-order tensor to recover missing entries, as discussed in Subsection \ref{katmtt}.
\item\emph{Image extraction}: Extracting a single recovered image from the recovered concatenated tensor, which is discussed in Subsection \ref{riextra}.
\end{enumerate}
\subsection{Concatenating images for tensor augmentation}\label{cifta}
Consider an $N$th-order color image tensor $\mc{X}\in\mathbb{R}^{I_1\times I_2\times 3}$ that consists of partially known entries given by a subset $\Omega$. Applying directly the KA scheme to $\mc{X}$, then subsequently the TMac-TT algorithm for completion, will result in blocking-artifacts as demonstrated in Fig. \ref{lenablocking}. To circumvent this problem, an initial preprocessing step prior to KA is added that concatenates identical copies of $\mc{X}$ to form a fourth-order tensor $\mc{X}_{ci}\in\mathbb{R}^{I_1\times I_2\times 3\times C}$, where $C>1$ is the number of copies of the tensor $\mc{X}$. In fact, \cite{bengua16} had naturally formed a tensor similar to $\mc{X}_{ci}$ for color video recovery, however, rather than have $C$ for the fourth mode, the label $T$ is used to represent the time frames of a color video. Therefore $\mc{X}_{ci}$ can be considered a motionless color video with $C$ frames.

\begin{figure}[!htbp]
\centering
\includegraphics[width=2.2in]{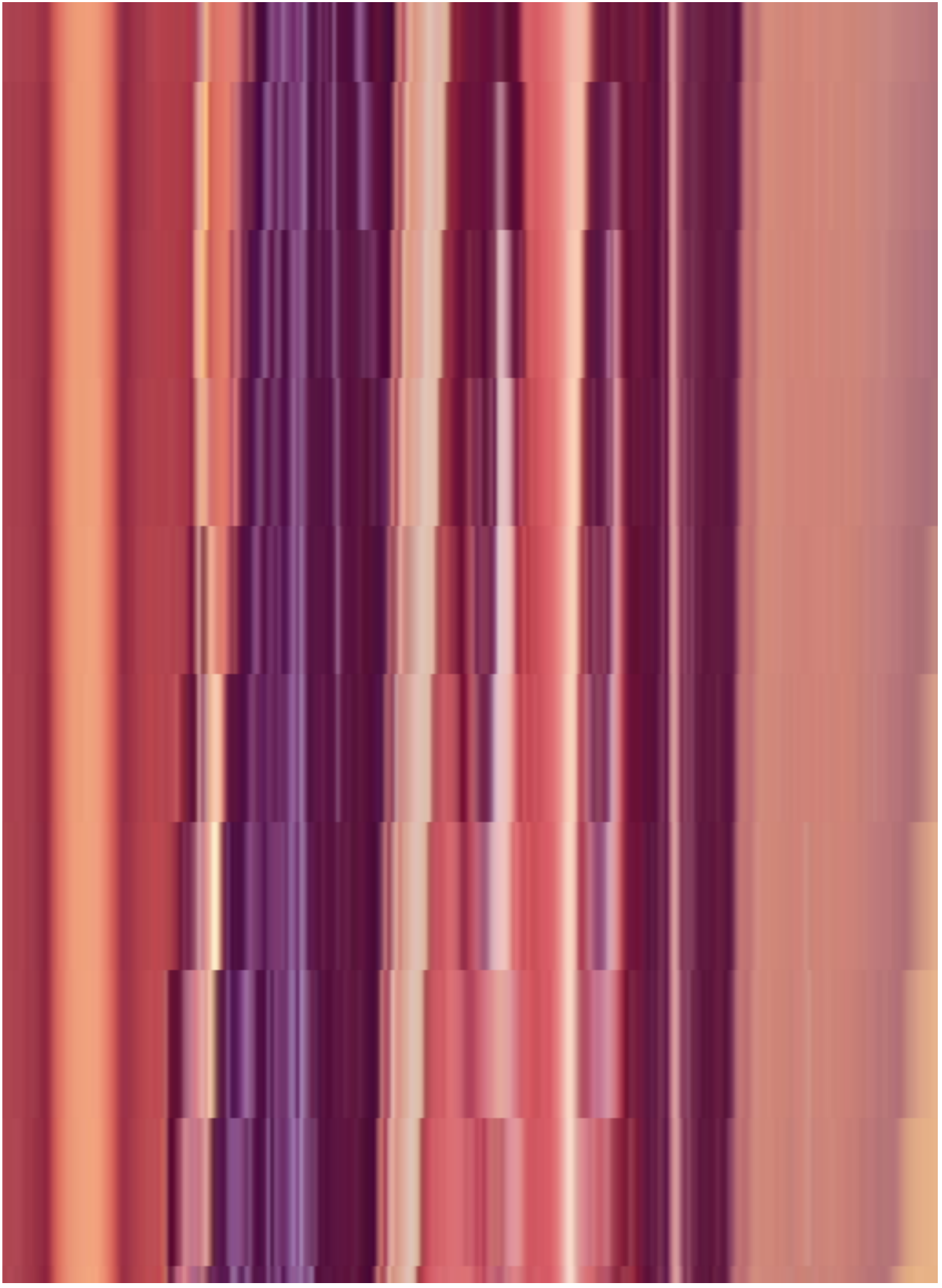}
\caption{An example of a third-order concatenated tensor $\mc{X}_{vst}$ of the Lena image.}\label{lenvstimg}
\end{figure}

The next step is to permute and reshape $\mc{X}_{ci}$ to a third-order tensor $\mc{X}_{vst}\in\mathbb{R}^{\tilde{I}_1\times I_2\times 3}$, where $\tilde{I}_1=CI_1$ is the \emph{combined row mode}. Displaying $\mc{X}_{vst}$ would result in a distorted continuous stream of the original image, hence, its structural properties have completely changed. The motivation to form this type of tensor is that the repetition of the image allows for more potential correlations, symmetry and/or continuity than a single image would exhibit. Additionally, applying KA on $\mc{X}_{vst}$ would result with obvious block-artifacts only on $\mc{X}_{vst}$, however, when the recovered image is extracted in the final step of ICTAC, these artifacts are minimized substantially.  Fig. \ref{lenvstimg} demonstrates $\mc{X}_{vst}$ for the original Lena image with no missing entries.
\begin{figure*}[!htbp]
\centering
\includegraphics[width=7in]{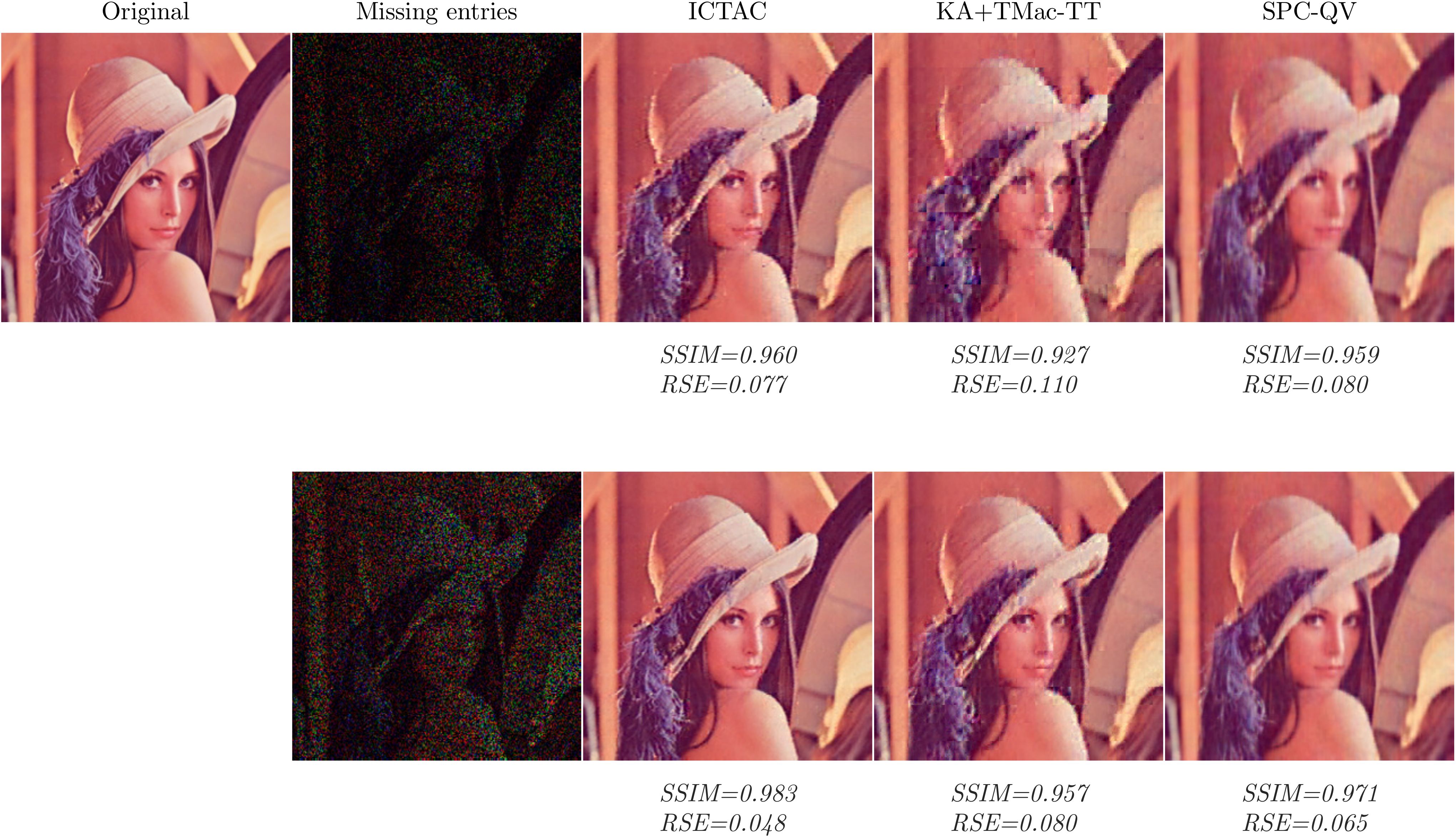}
\caption{Recovery of the Lena image for 90\% missing entries and 80\% missing entries. Top row from left to right: the original image, the original image with 90\% missing entries, and the subsequent recovery results for ICTAC, KA+TMac-TT and SPC-QV. Similarly for the bottom row from left to right: the original image with 80\% missing entries, then recovery results for ICTAC, KA+TMac-TT and SPC-QV.}\label{lenacomparecires}
\end{figure*}
We can see that there are considerably more patterns in the image compared to the original image itself. Therefore by converting an image to essentially a motionless video allows for new insights and potential for image completion tasks. The formation of a tensor of this type was originally defined as a \emph{video sequence tensor} (VST) in \cite{bengua16} for preprocessing a color video prior to KA+TMac-TT.

\subsection{KA and TMac-TT}\label{katmtt}
For the next step we apply the modified KA scheme in (\ref{moddedKA}) to augment the third-order tensor $\mc{X}_{vst}$ of size $\tilde{I}_1\times I_2\times 3$ to a higher-order tensor $\tilde{\mc{X}}\in\mathbb{R}^{I_1\times I_2\times\cdots\times I_K}$, with $K\geq N$. The modified KA scheme is needed due to subspace $\tilde{I}_1\times I_2$ of $\mc{X}_{vst}$ being a large non-square matrix, and the original KA scheme proposed in \cite{bengua16} only catered for tensors of the form $I_1\times I_2\times3$, with $I_1=I_2$, hence the subspace $I_1\times I_2$ is a square matrix.

After a block structured addressing via KA has been applied on $\mc{X}_{vst}$, then it is ready to be transferred to a tensor completion algorithm.  The TMac-TT algorithm can now be utilized to recover the missing entries of $\tilde{\mc{X}}$. For more detail on these steps, the reader can refer to \cite{bengua16}.

\subsection{Recovered image extraction}\label{riextra}
After $\tilde{\mc{X}}$ has been recovered via KA+TMac-TT, to recover a single image, an inverse KA scheme is utilized to obtain a third-order recovered concatenated tensor $\tilde{\mc{X}}_{vst}\in\mathbb{R}^{\tilde{I}_1\times I_2\times 3}$. $\tilde{\mc{X}}_{vst}$ is subsequently reshaped and permuted back to a fourth-order tensor $\tilde{\mc{X}}_{ci}\in\mathbb{R}^{I_1\times I_2\times 3\times C}$, and from this tensor we can extract a single recovered image, e.g. the image at $\tilde{\mc{X}}_{ci}(:,:,:,1)$.
\begin{figure*}[!htbp]
\centering
\includegraphics[width=7in]{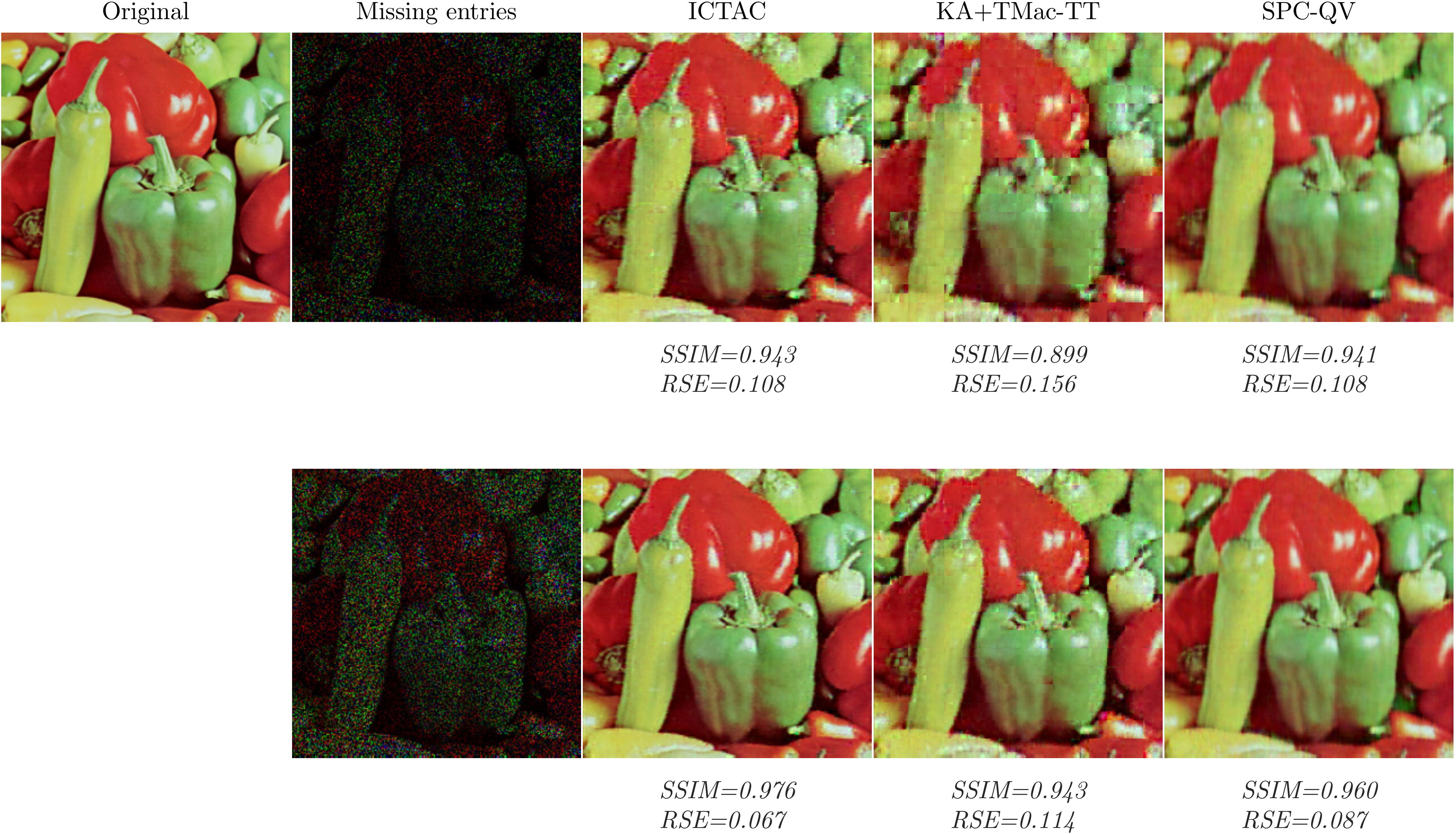}
\caption{Recovery of the Peppers image for 90\% missing entries and 80\% missing entries. Top row from left to right: the original image, the original image with 90\% missing entries, and the subsequent recovery results for ICTAC, KA+TMac-TT and SPC-QV. Similarly for the bottom row from left to right: the original image with 80\% missing entries, then recovery results for ICTAC, KA+TMac-TT and SPC-QV.}\label{peppcomparecires}
\end{figure*}
\section{Experimental results}\label{experres}
The experiments are conducted for image completion tasks of various missing ratios for the Lena and Peppers color images. The proposed ICTAC framework is benchmarked against current state-of-the-art tensor completion algorithms KA+TMac-TT \cite{bengua16} and SPC-QV \cite{YOKO15}.

The missing ratio ($mr$) as a percentage of a test image is defined as
\bea
mr = \frac{p}{\prod_{l=1}^{N}I_l}\times100\%,
\eea
with $p$ being the number of missing entries, which is chosen randomly based on a uniform distribution.

Performance measures include the relative square error (RSE) between an approximately recovered tensor and the original one, which is defined as,
\bea
RSE = ||\mc{X}-\mc{T}||_F/||\mc{T}||_F,
\label{RSE}
\eea
and the structural similarity index ($SSIM$) \cite{1284395}.

The algorithms ICTAC and SPC-QV recover color images represented by third-order tensors of size $243\times512\times3$ to cater for the modified KA scheme of ICTAC as discussed in the previous sections, whereas KA+TMac-TT uses images of size $256\times256\times3$ because the traditional KA scheme works only on the condition $I_1=I_2$. To compare the algorithms in a fair manner, the $RSE$ and $SSIM$ is calculated based on the same image sizes for the initial tensor with missing entries and the final recovered tensor, and there is no image distortion which can change the $mr$ throughout the runtime of the completion algorithms. The only distortion that may happen is after the algorithms have completed their calculations, where the final recovered tensors of size $243\times512\times3$ from ICTAC and SPC-QV is reshaped to $256\times256\times3$ so that a visual comparison can be made to the results of KA+TMacTT. The number of copied images $C=81$ for all benchmarks, and simulations are conducted using a Matlab environment. 

The ICTAC framework contains several tensor size transformations. For clarity, an outline of these changes is given for an initial tensor $\mc{X}\in\mathbb{R}^{243\times512\times3}$ to the higher-order tensor $\tilde{\mc{X}}$ in Subsection \ref{katmtt}: 
\begin{enumerate}
\item\emph{Image concatenation}: $\mc{X}\in\mathbb{R}^{243\times512\times 3}\to\mc{X}_{ci}\in\mathbb{R}^{243\times512\times 3\times81}$.
\item\emph{Obtaining a VST}: $\mc{X}_{ci}\in\mathbb{R}^{243\times512\times 3\times81}\to\mc{X}_{vst}\in\mathbb{R}^{19683\times512\times3}\nonumber$.
\item\emph{Applying the modified KA}: $\mc{X}_{vst}\in\mathbb{R}^{19683\times512\times3}\to\tilde{\mc{X}}\in\mathbb{R}^{6\times6\times6\times6\times6\times6\times6\times6\times6\times 3}$, i.e. $n=9$ in (\ref{moddedKA}).
\end{enumerate}
Fig. \ref{lenacomparecires} presents the results for the Lena completed images using ICTAC, KA+TMac-TT and SPC-QV. In the case of 90\% missing entries, the KA+TMac-TT algorithm performs the worst in terms of $RSE$ and $SSIM$ compared to the other two algorithms. This is a significant result because it was shown recently in \cite{bengua16} that KA+TMac-TT outperformed state-of-the-art algorithms in color image recovery. Comparing ICTAC and SPC-QV, it is interesting to see that although both algorithms had similar performance in $RSE$ and $SSIM$, with ICTAC slightly better in both, there are quite striking visual differences in their image recovery results. Specifically, SPC-QV tends to have a smoother uniform textures of the Lena image, especially around edges, however, a slight blur and fading can be seen that affects detail such as the hair strands when comparing it to the original image. The recovered ICTAC image demonstrates an attempt to detail the finer textures of the original image, which can be clearly seen on the hair strands, however, some block-artifacts and errors from the image recovery can be seen. For 80\% missing entries, ICTAC provides a recovered image almost completely similar to the original image with an SSIM of $0.983$ and RSE of $0.048$. SPC-QV has a slight image blur and still does not provide detail on more complicated parts of the image such as the hair and lips. The KA+TMac-TT results still has the worst performance with an RSE of $0.08$, and block-artifacts can still be easily observed.

Results for the Peppers image completion task is presented in Fig. \ref{peppcomparecires}. Similar to the Lena image results for 80\% and 90\% missing entries, KA+TMac-TT has the lowest $RSE$ and $SSIM$ for both cases, with obvious block-artifacts and less detail than ICTAC and SPC-QV. The results of SPC-QV is shown to still have blurring and fading effects, which decreases the detail of the image regardless of the $RSE$ or $SSIM$. ICTAC is shown to have slightly better $RSE$ and $SSIM$ against SPC-QV for both cases and attempts to recover fine details of the original image, but with some minimal block-artifacts that can still be observed.

\section{Conclusion}\label{conclusend}
In this paper, we proposed a novel framework known as concatenated image completion via tensor augmentation and completion (ICTAC). The framework formulates a tensor from a concatenation of identical copies of a single color image with missing entires, which provides additional patterns to support image completion algorithms. It then utilizes tensor augmentation based on modified KA, a TT rank-based tensor completion algorithm TMac-TT to impute the missing entries, and finally, an image extraction method to recover the completed image. Our method was shown to outperform recently proposed state-of-the-art tensor completion algorithms KA+TMac-TT and SPC-QV for color image completion tasks.

ICTAC is still at its infancy and there are many future endeavours to further improve the efficiency of the algorithm. A major direction is to investigate the concatenation of images in different tensor structures and copies $C$, as there may be a minimum or maximum $C$ that could determine how good a result we can obtain for tensor completion. The exploration of different KA schemes would also benefit the framework as well as other TT rank-based tensor completion algorithms.

\section*{Acknoledgement}
The authors would like to thank Tatsuya Yokoto and Andrzej Cichocki for generously providing their Matlab code of the SPC-QV algorithm.

\bibliographystyle{IEEEtran}
\bibliography{TC}
\end{document}